\documentclass{article}

\usepackage[preprint]{neurips_2022}

\usepackage[utf8]{inputenc} 
\usepackage[T1]{fontenc}    
\usepackage{hyperref}       
\usepackage{xr-hyper}
\usepackage{url}            
\usepackage{booktabs}       
\usepackage{amsfonts}       
\usepackage{nicefrac}       
\usepackage{microtype}      
\usepackage{xcolor}         
\usepackage{times}
\usepackage{graphicx}
\usepackage{amsmath}
\usepackage{algorithmic}
\usepackage{algorithm}
\usepackage{footnote}
\usepackage{multirow}
\usepackage{makecell}
\usepackage{bm}
\usepackage{amssymb}
\title{Unsupervised Deep Discriminant Analysis Based Clustering}

\author{%
  Jinyu Cai\footnotemark[1]\\
  Fuzhou University\\
  Shenzhen Research Institute of Big Data\\
  \texttt{jinyucai1995@gmail.com} \\
  \And
  Wenzhong Guo\footnotemark[1]\\
  Fuzhou University\\
  \texttt{guowenzhong@fzu.edu.cn} \\
  \AND
  Jicong Fan\footnotemark[2]\\
  The Chinese University of Hong Kong (Shenzhen)\\
  Shenzhen Research Institute of Big Data\\
  \texttt{fanjicong@cuhk.edu.cn} \\
}

\begin{document}

\maketitle
\renewcommand{\thefootnote}{\fnsymbol{footnote}} 
\footnotetext[1]{Contributed equally to this work.} 
\footnotetext[2]{Corresponding author.} 
\renewcommand{\thefootnote}{\arabic{footnote}}

\begin{abstract}
This work presents an unsupervised deep discriminant analysis for clustering. The method is based on deep neural networks and aims to minimize the intra-cluster discrepancy and maximize the inter-cluster discrepancy in an unsupervised manner. The method is able to project the data into a nonlinear low-dimensional latent space with compact and distinct distribution patterns such that the data clusters can be effectively identified. We further provide an extension of the method such that available graph information can be effectively exploited to improve the clustering performance. Extensive numerical results on image and non-image data with or without graph information demonstrate the effectiveness of the proposed methods.
\end{abstract}

\section{Introduction}
\label{sec1}
Clustering, which aims to aggregate similar data together and separate dissimilar data, is a fundamental yet challenging research problem in machine learning and data mining \citep{jain2010data}. Classicial clustering methods such as k-means clustering \citep{hartigan1979algorithm}, spectral clustering \citep{shi2000normalized,ng2001spectral}, and subspace clustering \citep{vidal2011subspace,LRR_PAMI_2013,matsushima2019selective,fan2021large} have been well studied but their performance are not satisfactory when the structures of data are quite complex or the features are not discriminative. Recently, many researchers \citep{xie2016unsupervised,yang2016joint,jiang2017variational,
ji2017deep,ghasedi2017deep,abavisani2020deep,tang2020unsupervised,
lv2021pseudo,ZhangDeep2021,peng2022xai} focus on improving classical clustering methods by exploiting the strengths of deep learning such as auto-encoder (AE) \citep{hinton2006reducing} in data representation. For instance, ~\citet{xie2016unsupervised} proposed a deep embedded clustering (DEC) method via integrating feature learning and clustering in a unified network.
\citet{ji2017deep} proposed a deep subspace clustering method based on AE and self-expression \citep{elhamifar2013sparse}. Note that those self-expression based deep clustering methods such as \citep{ji2017deep,ZhangDeep2021} have at least quadratic time and space complexity and hence are not applicable to very large datasets \citep{matsushima2019selective,fan2021large}.

More recently, a few researchers \citep{zhao2021graph,zhang2020adaptive,tu2021deep} proposed to integrate graph convolutional networks \citep{KipfW17,velivckovic2018graph} into deep clustering methods such that both Euclidean feature and graph structure can be utilized for clustering. For instance, \citet{bo2020structural} proposed a structural deep clustering network (SDCN), which uses GCN module to provide graph structure information among data in the clustering-oriented embedding learning. \citet{peng2021attention} proposed an attention graph clustering network (AGCN), which improves SDCN by introducing an attention mechanism to dynamically fuse the embedded representation and graph feature, as well as the representations learned by different GCN layers. In these methods, it is expected that exploiting more useful information can improve the clustering performance.

The aforementioned deep clustering methods do not explicitly ensure distinct and compact cluster structures, which may lead to difficulties in partitioning data points around decision boundaries.  An intuitive example is shown by Figure \ref{visualization} (also Figure \ref{DDAC-visualization} in the supplement). In this paper, we propose a novel clustering method called deep discriminative analysis based clustering (DDAC).
Specifically, we use an autoencoder to learn the embedded representation for the input data and propose to minimize the intra-cluster discrepancy and maximize the inter-cluster discrepancy of the data representation to pursue more discriminative and compact latent representation. We also provide an extension of DDAC that is able to effectively exploit the graph information for clustering.
Our contributions are as follows.

\begin{itemize}
\setlength{\itemsep}{0pt}
\setlength{\parsep}{0pt}
\setlength{\parskip}{0pt}
\item We present a novel unsupervised deep discriminant analysis based clustering method, which extends linear discriminative analysis to unsupervised deep learning.
\item We propose an effective regularizer to obtain small intra-cluster discrepancy and large inter-cluster discrepancy in the data.
To ensure unsupervised discriminative projection,  we provide a dynamic confidence assignment mechanism to increase the reliability of the intra-cluster discrepancy.
\item We further extend our method to graph based clustering. It enables us to take advantages of both feature information and graph information to do clustering in an unsupervised deep discriminative manner.
\end{itemize}

Empirical studies on many benchmark datasets show that our methods can outperform the baselines.


\section{Deep Discriminant Analysis Based Clustering (DDAC)}
\label{sec2.2}
Suppose we have a data matrix $\mathbf{X}\in \mathbb{R}^{n \times d}$, where $n$ denotes the number of samples and $d$ denotes the number of features or variables. Our goal is to cluster the rows of $\mathbf{X}$ into $k$ groups. We consider the case that the variables of $\mathbf{X}$ are not discriminative such that classical clustering methods such as k-means, spectral clustering, and subspace clustering are not effective in identifying the clusters. Therefore, instead of directly clustering the rows of $\mathbf{X}$, we want to extract some discriminative features  denoted by $\mathbf{Z}=[\mathbf{z}_1,\mathbf{z}_2,\ldots \mathbf{z}_n]^\top\in\mathbb{R}^{n\times d'}$ from $\mathbf{X}$ and then cluster the rows of $\mathbf{Z}$. Specifically, we use a feature extractor $h_{\theta}$ to get
\begin{equation}
\mathbf{z}_i=h_{\theta}(\mathbf{x}_i),\quad i=1,2,\ldots,n,
\end{equation}
where $\theta$ denotes the set of parameters that can be learned from $\mathbf{X}$.
Suppose we have already known the labels of the columns of $\mathbf{Z}$.  Then we define an intra-cluster discrepancy as
\begin{eqnarray}
\label{eqintra1}
D_{\text{intra}}:= \frac{1}{n}{ \sum_{j=1}^{k}  \sum_{i\in c_j} \left \| \textbf{z}_{i}-{\bm{\mu}_{j}} \right \| ^{2}},
\end{eqnarray}
where $\bm{\mu}_j=\frac{1}{\vert c_j\vert}\sum_{i\in c_j} \mathbf{z}_i$ is the centroid of cluster $j$ and $c_j$ denotes the set of indices of data points in cluster $j$. Besides, we also define an inter-cluster discrepancy as
\begin{eqnarray}
\label{eqinter1}
D_{\text{inter}}:=\sum_{j=1}^{k}\sum_{m=1}^{k} \left \| \bm{\mu}_{j}-\bm{\mu}_{m} \right \|^{2}.
\end{eqnarray}
We say that $\mathbf{Z}$ is discriminative if
\begin{equation}
\mathcal{L}_{\textrm{disc}}:={D_{\text{intra}}}/{D_{\text{inter}}}
\end{equation}
is large. We hope to learn a $h_{\theta}$ to maximize $\mathcal{L}_{\textrm{disc}}$ as much as possible. This is similar to the idea of Fisher linear discriminative analysis (LDA) \citep{bishop2006pattern} that aims to learn a linear projection to maximize the between-class variance and minimize the within-class variance.

However, our goal is clustering and hence the labels of $\mathbf{Z}$ are unknown.  First, we need to ensure that $\mathbf{Z}$ should preserve as much information of $\mathbf{X}$ as possible. Therefore, we propose to use an encoder-decoder model to obtain $\mathbf{Z}$, i.e., minimizing
\begin{equation}
\begin{aligned}
\mathcal{L}_{\text{recon}} :=& \frac{1}{2n}\sum_{i=1}^{n} \left \| \mathbf{x}_{i}- f_{\theta'}({\mathbf{z}_{i}}) \right \| ^{2}=\frac{1}{2n}\sum_{i=1}^{n} \left \| \mathbf{x}_{i}- f_{\theta'}({h_{\theta}(\mathbf{x}_i)}) \right \| ^{2},
\end{aligned}
\label{eq:ReconstructionLoss}
\end{equation}
where $f_{\theta'}(\cdot)$ denotes the decoder with parameters set $\theta'$. Both $h_{\theta}$ and $f_{\theta'}$ are neural networks.

Now we propose to estimate the labels of $\mathbf{Z}$ iteratively.
Specifically, following the definition in ~\citep{van2008visualizing,xie2016unsupervised}, we define a matrix $\mathbf{Q}\in\mathbb{R}^{n\times k}$ as
\begin{eqnarray}
\label{eqsoft}
q_{ij}=\frac{(1+\left \| \mathbf{z}_{i}-\bm{\mu} _{j} \right \|^{2})^{-1}}{\sum_{j'=1}^{k} (1+\left \| \mathbf{z}_{i}-\bm{\mu} _{j'} \right \|^{2})^{-1}},\quad i=1,\ldots,n,~j=1,\ldots,k,
\end{eqnarray}
where ${\bm{\mu}}_{j}$ denotes the $j$-th cluster centroid and is initialized by $k$-means performed on a temporary $\mathbf{Z}$. $q_{ij}$ denotes the probability that allocates sample $i$ to cluster $j$. Therefore, $\mathbf{Q}$ is a soft label matrix but can be far from the true label matrix. Similar to \citep{xie2016unsupervised}, we introduce a reinforced soft label matrix $\mathbf{P}$ by
\begin{eqnarray}
\label{eqtarget}
p_{ij}=\frac{q_{ij}^{2}/\sum _{i=1}^{n}q_{ij}}{\sum _{j'=1}^{k}q_{ij'}^{2}/\sum _{i=1}^{n}q_{ij'}}, \quad i=1,\ldots,n,~j=1,\ldots,k.
\end{eqnarray}
$\mathbf{P}$ should be sparser than $\mathbf{Q}$ and the very large or very small values in $\mathbf{P}$ are more reliable than those in $\mathbf{Q}$ as soft labels.
Nevertheless, $\mathbf{P}$ or $\mathbf{Q}$  cannot be directly used to generate labels for $\mathbf{Z}$ when considering \eqref{eqintra1}. The reason is that the probabilities given by \eqref{eqsoft} are just estimations of the clustering tendency and the unreliable probabilities can lead to unreliable discriminative latent representation.  Therefore, we design a  confident assignment selection mechanism to mitigate the negative effects of unreliable probabilities. It is mathematically formulated as an indicator vector $\mathbf{t}\in\mathbb{R}^n$ defined by
\begin{align}
\label{eqthreshold}
t_{i} = \left\{\begin{matrix}
1, & \max_{1\leq j\leq k}(p_{ij}) > \delta,\\
0, & \mbox{ otherwise. }
\end{matrix}\right.
\end{align}
Here $\delta$ denotes a global threshold above which the clustering probability for each sample is confident.

Now based on \eqref{eqintra1}, \eqref{eqtarget}, and  \eqref{eqthreshold},  we can define a reliable estimate of the intra-cluster discrepancy as
\begin{eqnarray}
\label{eqintra2}
\hat{D}_{\textrm{intra}}:= \frac{ \sum_{j=1}^{k}  \sum_{i=1}^{n} \left \| \textbf{z}_{i}-\bm{\mu}_{j} \right \| ^{2}\times p_{ij} \times t_{i}}{\sum_{i=1}^{n} t_{i}}.
\end{eqnarray}
An estimate of the inter-cluster discrepancy is given by
$\hat{D}_{\textrm{inter}}:=\sum_{j=1}^{k}\sum_{m=1}^{k} \left \| \bm{\mu}_{m}-\bm{\mu}_{j} \right \|^{2}={D}_{\textrm{inter}}$.
Now we minimize
\begin{eqnarray}
\displaystyle
\label{eqdisc}
\mathcal{L}_{\textrm{disc}}:=\frac{\hat{D}_{\textrm{intra}}}{\hat{D}_{\textrm{inter}}}= \frac{\sum_{j=1}^{k}  \sum_{i=1}^{n} \left \| \textbf{z}_{i}-\bm{\mu}_{j} \right \| ^{2}\times p_{ij} \times t_{i} }{(\sum_{i=1}^{n} t_{i}) \times \sum_{j=1}^{k}  \sum_{m=1}^{k}  \left \| \bm{\mu}_{m}-\bm{\mu}_{j} \right \|^{2}}
\end{eqnarray}
to obtain a reliable discriminative representation.

Note that in LDA,  the variables in the latent space are uncorrelated because the projection vectors are orthogonal.  The uncorrelated features are more useful than correlated features. Hence,  in our method, we propose to make the columns of $\mathbf{Z}$ orthogonal via considering the following regularization term
\begin{eqnarray}
\label{eqorth}
\mathcal{L}_{\textrm{orth}}:=\frac{1}{\sum_{i=1}^{n} t_{i}}\left \| \hat{\mathbf{Z}}^{\top}\hat{\mathbf{Z}}\odot (\mathbf{O- \hat{\mathbf{I}}}) \right \|_{F}^{2},
\end{eqnarray}
where $\odot$ denotes Hadamard product, $\hat{\mathbf{Z}}=\mathbf{Z} \odot \mathbf{1}^{\top}\mathbf{t}$ is the selective representation from the confident assignment. $\mathbf{O} \in \mathbb{R}^{d' \times d'}$ is a matrix with all elements are 1 and $\hat{\mathbf{I}}\in \mathbb{R}^{d' \times d'}$ is an identity matrix.

In addition, let $\Omega$ be the index set of samples satisfying the condition of confident assignment selection and we want $\mathbf{Q}_{\Omega}$ to be close to $\mathbf{P}_{\Omega}$ via minimizing
\begin{align}
\label{eqlc}
\mathcal{L}_{\textrm{clus}} &:= KL(\mathbf{P}_{\Omega}\left |  \right |\mathbf{Q}_{\Omega})
      = \sum _{i \in \Omega }\sum _{j=1}^{k} p_{ij}\log\frac{p_{ij}}{q_{ij}}
      = \sum _{i=1}^{n} t_{i}\sum _{j=1}^{k} p_{ij}\log\frac{p_{ij}}{q_{ij}}.
\end{align}

Finally, we have an overall objective function
\begin{eqnarray}\label{eq_all}
\mathcal{L}( \Theta,   \left \{ \bm{\mu} \right \}):= \mathcal{L}_{\textrm{recon}}+ \alpha \mathcal{L}_{\textrm{clus}}+\beta \mathcal{L}_{\textrm{disc}}+ \gamma \mathcal{L}_{\textrm{orth}},
\end{eqnarray}
where $\Theta= \left \{ \theta, \theta' \right \}$ and $\left \{ \bm{\mu} \right \} = \left \{ \bm{\mu}_{1}, \cdots, \bm{\mu}_{k} \right \} $ are the network parameters  and cluster centroids respectively. $\alpha$, $\beta$ and $\gamma$ are hyper-parameters.

We summarize the roles of all terms of \eqref{eq_all} as follows:
\begin{itemize}
\setlength{\itemsep}{0pt}
\setlength{\parsep}{0pt}
\setlength{\parskip}{0pt}
\item $\mathcal{L}_{\textrm{recon}}$ aims to preserve the inherent structure of the input data.
\item $\mathcal{L}_{\textrm{clus}}$ ensures the network to learn clustering assignment from the embedded representation $\mathbf{Z}$ of the input data.
\item $\mathcal{L}_{\textrm{disc}}$ tries to minimize of intra-cluster discrepancy and maximize the inter-cluster discrepancy in the latent space, thereby making the learned representation more discriminative.
\item $\mathcal{L}_{\textrm{orth}}$ aims to make the columns of $\mathbf{Z}$ orthogonal so as to obtain uncorrelated latent features.
\end{itemize}

\begin{figure}[t]
\centering
\includegraphics[width=5.5in]{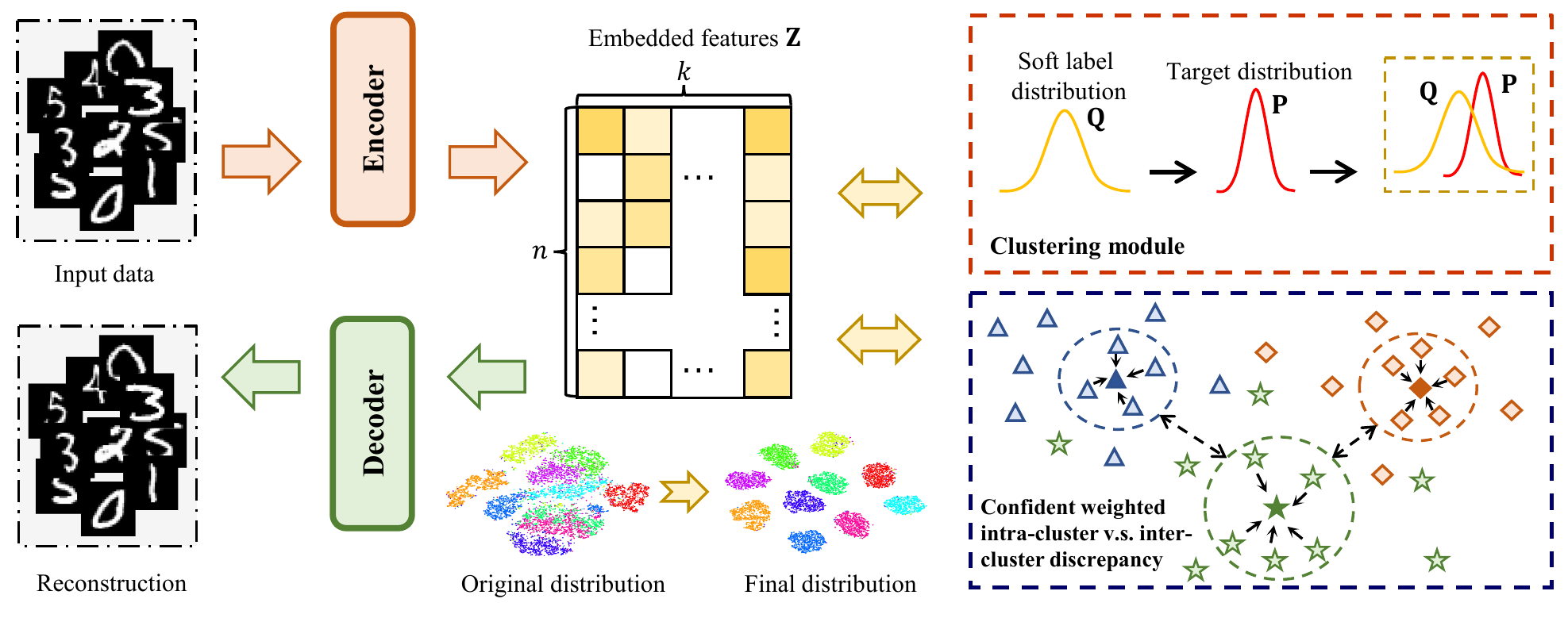}\\
\caption{The network architecture of the proposed DDAC method. The input data is first encoded into a low-dimensional embedding features $\mathbf{Z}$ and then $\mathbf{Z}$ is decoded to reconstruct the input data. In the embedding layer, a clustering module is introduced to produce cluster label assignment with $\mathbf{Z}$. Meanwhile, we propose to minimize the intra-cluster discrepancy and maximize the inter-cluster discrepancy of the data representation to pursue discriminative and compact latent representation.}
\label{DDACNetwork}
\end{figure}

Figure~\ref{DDACNetwork} shows the network architecture of the proposed method.  The training procedure of the proposed method is illustrated in Algorithm~\ref{Algorithm}, where the final cluster labels are obtained from the soft label distribution $\mathbf{Q}$.
\begin{algorithm}[h]
\caption{Deep Discriminative Analysis Based Clustering (DDAC)}
\label{Algorithm}
\textbf{Input}: Data matrix $\mathbf{X}$, dimension of latent space $d'$, number of clusters $k$, hyper-parameters $\alpha$, $\beta$ and $\gamma$, threshold of confident assignment $\delta$, total training iterations $T$.\\
\textbf{Output}: Cluster labels $C$.

\begin{algorithmic}[1]
\STATE {Initialize $\Theta$, pretrain the autoencoder, and get $\mathbf{Z}$}
\STATE {Initialize the cluster centroids $\{\bm{\mu}\}$ with $k$-means on $\mathbf{Z}$}
      \FOR{$iter=1$ to $T$}
      \STATE {Compute the soft label distribution $\mathbf{Q}$ using Eq.~\eqref{eqsoft}}
      \STATE {Compute the target distribution $\mathbf{P}$ using Eq.~\eqref{eqtarget}}
      \STATE {Obtain the selective index set $\mathbf{t}$ using Eq.~\eqref{eqthreshold}}
      \STATE {Calculate $\mathcal{L}_{\textrm{recon}}$, $\mathcal{L}_{\textrm{clus}}$, $\mathcal{L}_{\textrm{disc}}$, and $\mathcal{L}_{\textrm{orth}}$ using Eqs.~\eqref{eq:ReconstructionLoss},~\eqref{eqlc},~\eqref{eqdisc}, and \eqref{eqorth} respectively}
      \STATE {Update $\Theta$ and $ \left \{ \bm{\mu} \right \}$ by minimizing Eq.~\eqref{eq_all} with some optimizer and get $\mathbf{Z}$}
   \ENDFOR
\STATE {Obtain the final clustering results as $c_{i} = \mathop{\arg\max}_j~ q_{ij}$}
\RETURN {$C$.}
\end{algorithmic}
\end{algorithm}

\section{DDAC with Graph Information}

Suppose besides the data matrix $\mathbf{X}\in \mathbb{R}^{n \times d}$,  we also have an associated (undirected) graph $G=(V,E)$,  in which $V=\{1,2,\ldots,n\}$ denotes the set of vertices corresponding to the data points of $\mathbf{X}$ and $E=\{e_{ij}\}$ denotes the (possibly weighted) edges between the data points. Or equivalently, we can replaced $E$ by an adjacency matrix $\mathbf{A}\in \mathbb{R}^{n \times n}$, in which ${a}_{ij}$ denotes the similarity between vertex $i$ and vertex $j$.  Our goal is to cluster the rows of $\mathbf{X}$ or partition the nodes of $G$ into $k$ groups. Note that if $\mathbf{A}$ is not given, we can generate it from $\mathbf{X}$ via some approaches such as $k$ nearest neighbors ($k$NN).

In this section, we propose an extension of DDAC to take advantages of the graph structure. To this end, we incorporate GCN~\citep{KipfW17} into our DDAC to facilitate clustering, yielding a method called DDAC-G. The network architecture of DDAC-G is presented in Appendix~\ref{A2} due to the limitation of paper length. For a GCN module with $L$ layers, we can formulate the learned latent features $\mathbf{S}_{l}$ in the $l$-th layer as:
\begin{eqnarray}
\label{eqgcnz}
\mathbf{S}_{l}= \sigma ( \tilde{\mathbf{D}}^{-\frac{1}{2}} \tilde{\mathbf{A}}\tilde{\mathbf{D}}^{-\frac{1}{2}} \tilde{\mathbf{S}}_{l-1} \mathbf{W}_{l} ),
\end{eqnarray}
where $\mathbf{W}_{l}$ is the weight matrix of the $l$-th layer of the GCN and $\sigma(\cdot)$ denotes the activation function (e.g. ReLU). $\tilde{\mathbf{A}} = \mathbf{A} + \mathbf{I} $,  where $\mathbf{I}\in \mathbb{R}^{n\times n}$ is an identity matrix.  $ \tilde{\mathbf{D}}$ is the degree matrix of $\tilde{\mathbf{A}}$. In order to integrate both the representations learned by AE and GCN, we let $\tilde{\mathbf{S}}_{l-1} = (1-\epsilon) \mathbf{S}_{l-1} + \epsilon \mathbf{Z}_{l-1} $, where $\epsilon$ is fixed as 0.5.  Particularly, the input of the first layer of GCN is the original data $\mathbf{X}$, i.e.,
\begin{eqnarray}
\label{eqgcnz1}
\mathbf{S}_{1}= \sigma ( \tilde{\mathbf{D}}^{-\frac{1}{2}} \tilde{\mathbf{A}}\tilde{\mathbf{D}}^{-\frac{1}{2}} \mathbf{X}\mathbf{W}_{1} ).
\end{eqnarray}

The last layer of the GCN utilizes the $\textrm{softmax}$ function to obtain a probability distribution matrix $\mathbf{Y}$ for the categorization of the input data:
\begin{eqnarray}
\label{eqgcnzoutput}
\mathbf{Y} = \textrm{softmax}( \tilde{\mathbf{D}}^{-\frac{1}{2}} \tilde{\mathbf{A}}\tilde{\mathbf{D}}^{-\frac{1}{2}} \mathbf{S}_{L-1}\mathbf{W}_{L} ),
\end{eqnarray}
where $y_{ij}$ represents the probability that the GCN network predicts a sample $i$ into cluster $j$. Consequently, by considering both the predictive distribution of AE and GCN, the clustering loss of the DDAC-G method is
\begin{align}
\label{eqlc-g}
\mathcal{L}_{\textrm{g-clus}} &:= \alpha_{1} KL(\mathbf{P}_{\Omega}\left |  \right |\mathbf{Q}_{\Omega}) + \alpha_{2} KL(\mathbf{P}_{\Omega}\left |  \right |\mathbf{Y}_{\Omega})\nonumber\\
      &=\alpha_{1} \sum _{i \in \Omega }\sum _{j=1}^{k} p_{ij}\log\frac{p_{ij}}{q_{ij}} + \alpha_{2} \sum _{i \in \Omega}\sum _{j=1}^{k} p_{ij}\log\frac{p_{ij}}{y_{ij}}\nonumber\\
      &= \alpha_{1} \sum _{i=1}^{n} t_{i}\sum _{j=1}^{k} p_{ij}\log\frac{p_{ij}}{q_{ij}} + \alpha_{2} \sum _{i=1}^{n} t_{i}\sum _{j=1}^{k} p_{ij}\log\frac{p_{ij}}{y_{ij}},
\end{align}
where $\Omega$ denotes the index set of samples that satisfy the condition of confident assignment selection. $\alpha_{1}$ and $\alpha_{2}$ control weights of the two modules and are set to 0.1 and 0.01 respectively in this paper.

Now, based on \eqref{eq:ReconstructionLoss}, \eqref{eqdisc}, \eqref{eqorth} and \eqref{eqlc-g}, the overall objective function of the DDAC-G method is given as
\begin{eqnarray}\label{eq_all_ddgc}
\mathcal{L}( \Theta,   \left \{ \bm{\mu} \right \}, \mathcal{W}):= \mathcal{L}_{\textrm{recon}}+\mathcal{L}_{\textrm{g-clus}}+\beta \mathcal{L}_{\textrm{disc}}+ \gamma \mathcal{L}_{\textrm{orth}},
\end{eqnarray}
where $\Theta= \left \{ \theta, \theta' \right \}$, $\left \{ \bm{\mu} \right \} = \left \{ \bm{\mu}_{1}, \cdots, \bm{\mu}_{k} \right \} $, and $\mathcal{W}={\left \{ \mathbf{W}_{1}, \cdots, \mathbf{W}_{L} \right \} }$ are the AE parameters, cluster centroids, and GCN parameters, respectively. $\beta$ and $\gamma$ are two hyper-parameters. It is worth noting that we integrate the learned features in each layer of GCN and AE to obtain more informative features, i.e., $\tilde{\mathbf{S}}$, as the input to the subsequent layer of GCN. Therefore, it makes sense to use the GCN prediction $\mathbf{Y}$ rather than $\mathbf{Q}$ to obtain the cluster assignment. Specifically, $c_i=\mathop{\arg\max}_j~y_{ij}$.

\section{Connection with previous work}

The main idea of DDAC and DDAC-G is motivated from LDA \citep{bishop2006pattern}. However,  LDA is a supervised learning method and requires reliable labels, which are difficult to obtain in unsupervised scenarios especially when we are using the latent representations of deep neural networks. Therefore, it is non-trivial to extend LDA to deep learning and clustering.

Note that \citet{ding2007adaptive} proposed to perform $k$-means and LDA alternately, which showed much better clustering performance compared to vanilla $k$-means. \citet{ye2007discriminative} showed that the LDA projection can be factored out from the integrated LDA subspace selection and clustering formulation and proposed an algorithm called discriminative k-means for simultaneous LDA subspace selection and clustering.
\citet{8489417} introduced a discriminative loss to deep clustering. The discriminative loss can be regarded as a graph regularization for the latent representations, where the similarities are the predicted probability of softmax functions. However, the clustering accuracy (Table I in \citep{8489417}) of the method is not high compared to the baselines used in our paper. \citet{tzoreff2018deep} proposed to obtain discriminative latent space in the pre-training (AE) phase, where the discriminative information are from k-nearest neighbors of data space. The clustering accuracy (Table 2 in \citep{tzoreff2018deep}) is lower than our method (Table 1 in our paper).
\citet{chang2019deep} proposed a method called deep discriminative clustering that
models the clustering task by investigating relationships between patterns with a deep neural network. They introduced a global constraint to adaptively estimate the
relationships, and a local constraint based on cluster indicators to learn high-level discriminative representations. The strategy of pursuing discriminative features is different from ours.


\section{Experiments on data without graph information}
\subsection{Experimental settings}
We demonstrate the effectiveness of the proposed DDAC in comparison to sixteen baselines (including more than ten strong competitors) on two benchmark datasets MNIST\footnote{http://yann.lecun.com/exdb/mnist/} and Fashion-MNIST\footnote{https://www.kaggle.com/zalando-research/fashionmnist}, which both have 70,000 images of size $28\times 28$ belonging to 10 classes.
In our DDAC, for both datasets, we set $\alpha=0.1$ and choose Adam~\citep{kingma2015adam} as the optimizer, where the learning rate is 0.001. Since the two datasets are image datasets, we construct our DDAC with fully connected network (FCN) and convolutional neural network (CNN), yielding two methods DDAC-FCN and DDAC-CNN. Specifically, for DDAC-FCN, we use a $d$-500-500-1000-$d'$ fully-connected encoder and a symmetric decoder on MNIST and the $d$-500-500-2000-$d'$ ones on Fashion-MNIST, where $d'$ is set to 10. For DDAC-CNN, we uniformly use $3 \times 3$ kernel and convolutional auto-encoder whose encoder and decoder consist of four convolutional layers and two linear layers respectively. The more detailed settings of the network architectures are described in Appendix~\ref{A2} of the supplementary  material. We first pre-train an auto-encoder with the same network structure as our method for 50 epochs of batch-size 512, and initialize the cluster centroids with $k$-means of 20 runs. Then we train our methods for 200 epochs to obtain the cluster assignment. We use three popular metrics including Clustering Accuracy (ACC), Normalized Mutual Information (NMI), and Adjusted Rand Index (ARI) to evaluate the clustering performance.

\subsection{Comparative results}
We report the clustering results in terms of three evaluation metrics in  Table~\ref{DDAC-Clusteringresults}. We see that the proposed methods DDAC-FCN and DDAC-CNN outperform other methods in most cases. It is worth noting that the effectiveness of the proposed discriminative regularizer is demonstrated when we compare DDAC-FCN with DEC and IDEC. The superiority of DDAC-FCN over DEC and IDEC stems from the fact that the large intra-cluster discrepancy and small inter-cluster discrepancy provided by DDAC makes the learned representations more discriminative. Moreover, compared to several state-of-the-art methods, DDAC-FCN and DDAC-CNN also show competitive performance. For instance, in comparison to the second best results (e.g. SENet, NCSC and DEPICT), DDAC-CNN achieves 1.30\%, 3.07\%, 2.62\% improvement in ACC, NMI and ARI on MNIST, as well as 1.66\% improvement in NMI on Fashion-MNIST compared to the second best results.

\begin{table*}[h]
\centering
\caption{Clustering performance on MNIST and Fashion-MNIST.  The best three results are highlighted in bold. $h_{\max}$ denotes the size of the largest hidden layer of neural network. $h'$ denotes the size of the layer connected with the largest hidden layer of neural network. $d$ denotes the dimension of the data or the latent representation. $\omega$, $F$, and $\kappa$ denote the feature map size, kernel size, and channel number of CNN respectively.}
\resizebox{\textwidth}{!}{
\renewcommand{\arraystretch}{1.15}
\begin{tabular}{l|ccc|ccc|c}

\toprule
\multirow{2}{*}{Methods/Datasets} & \multicolumn{3}{c|}{MNIST} & \multicolumn{3}{c|}{Fashion-MNIST}  &Time \\
\cmidrule(r){2-4} \cmidrule(r){5-7}
&  ACC     &  NMI &  ARI
&  ACC     &  NMI &  ARI   &Complexity   \\

\midrule
$k$-means~\citep{macqueen1967some}     &54.10 & 50.70 &36.70   & 50.50  & 57.80  &40.30    & $O(nk)$   \\

SC~\citep{ng2001spectral}  &69.60 &66.30 &52.10   &50.80 &57.50 &--  & $O(n^{2}k)$\\

AE~\citep{hinton2006reducing}        &78.53 & 74.90 &71.64     & 56.72  & 55.35  &41.29   & $O(nh_{max}h')$      \\

DEC~\citep{xie2016unsupervised}        &86.53  &83.69 &80.29    & 57.81  &62.83 &45.71   &$O(nh_{max}h'+ndk)$      \\

JULE~\citep{yang2016joint}      &96.40 & 91.30  &92.70    & 56.30  & 60.80  & 39.60    &     \slash \\

DCN~\citep{yang2017towards} &85.47 &81.73 &76.26      &53.87 &58.84 &40.84  &  \slash \\

IDEC~\citep{guo2017improved}       &88.01  &86.38  &83.25      &57.64   &60.13 &44.09 & $O(nh_{max}h'+ndk)$     \\

VaDE ~\citep{jiang2017variational}      &94.50 & 87.60   &91.30   & 55.20  & 57.30   &42.30    &  $O(nh_{max}h'+ndk)$  \\

DEPICT~\citep{ghasedi2017deep}     &96.50 & 91.70  &\bf93.20    & 39.20 & 39.20   &30.40   & \slash      \\

ClusterGAN~\citep{mukherjee2019clustergan} &95.00 &89.00  &89.00     & 63.00  & 64.00     &50.00 &  \slash   \\

ASPC(w/o DA)~\citep{guo2019adaptive}  &93.10 &88.60 &--  &59.10 & 65.40 &-- & \slash\\

NCSC~\citep{zhang2019neural} &94.09 &86.12 &87.52 &\bf72.14 &\bf68.60 &\bf59.17 & $O(dn^2)$\\

DFCM~\citep{feng2020deep}  &88.17 &86.54 &83.37 &62.29 &64.54 &48.65  &\slash \\

$k$-DAE~\citep{opochinsky2020k} &88.00 &86.00 &82.00    &60.00 &65.00 &48.00 &$O(kh_{\max}h'n)$\\

SENet~\citep{zhang2021learning} &\bf96.80 &\bf91.80 &93.10 &\bf69.70 &66.30 &\bf55.60 & $O(d\rho n^2)$, $0<\rho<1$\\

TELL~\citep{peng2022xai} &95.16  &88.83  &89.66 &52.66&62.05 &43.35 & \slash \\

\midrule
DDAC-FCN    &\bf97.12 &\bf92.86   &\bf93.74    &67.52 &\bf69.08 &55.36  &    $O(nh_{\max}h'+ndk)$    \\

DDAC-CNN    &\bf98.10 &\bf94.87   &\bf95.82    &\bf68.69 &\bf70.26 &\bf57.39     &    $O(\omega^2F^2\kappa n+ndk)$ \\
\bottomrule
\end{tabular}}
\label{DDAC-Clusteringresults}
\end{table*}

It should be pointed out that although some subspace clustering methods such as NCSC and SENet have good clustering performance, they have quadratic (in terms of the number of data points) time and space complexities and hence do not scale to large datasets. In contrast, our DDAC methods have linear time and space complexities and are applicable to very large datasets. For completeness, we show the time complexities of some baselines in Table~\ref{DDAC-Clusteringresults}.

\begin{figure}[h]
\centering
\includegraphics[width=5.54in]{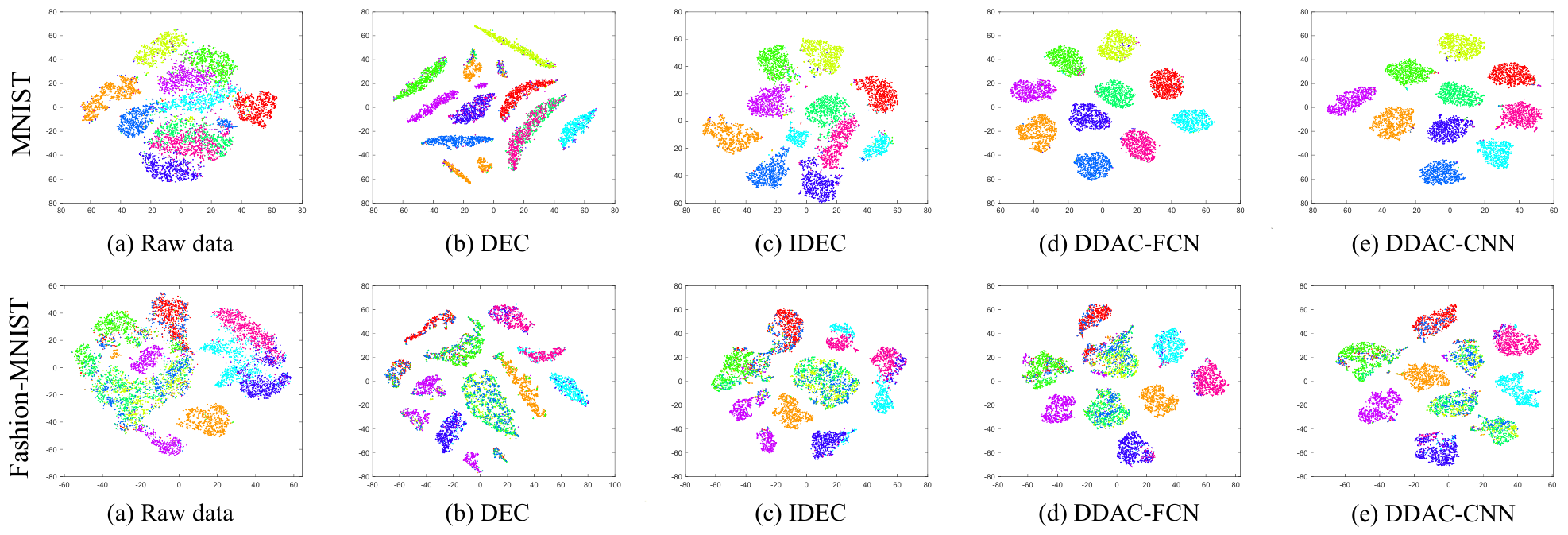}\\
\caption{The comparison of different methods on the t-SNE visualization of the learned representations. The first row is the visualization for MNIST, and the second row is for Fashion-MNIST.}
\label{DDAC-visualization}
\end{figure}

We further use t-SNE~\citep{van2008visualizing} to provide a visual comparison of clustering performance. Shown in Figure~\ref{DDAC-visualization}, on MNIST, our methods (DDAC-FCN and DDAC-CNN) have much more compact inter-class structures and distinct inter-class discrepancy than DEC and IDEC. Although clustering on Fashion-MNIST is difficult, our methods still significantly outperform DEC and IDEC visually. By the way, DDAC-CNN performs better than DDAC-FCN in this study.

\section{Experiments on data with graph information}
\label{sec3}
\subsection{Experimental settings}
We evaluate the proposed method DDAC-G on clustering data with graph information. We consider three non-graph datasets (USPS\footnote{https://www.csie.ntu.edu.tw/~cjlin/libsvmtools/datasets/}, REUTERS-10K\footnote{https://keras.io/api/datasets/reuters/}, HHAR\footnote{https://archive.ics.uci.edu/ml/datasets/Human+Activity+Recognition+Using+Smartphones}) and three graph datasets (DBLP\footnote{https://dblp.uni-trier.de}, CITESEER\footnote{http://citeseerx.ist.psu.edu/index}, and ACM\footnote{http://dl.acm.org/}), which are detailed in Table~\ref{Dataset2}. Note that for the non-graph datasets, we use $k$NN to construct graphs. More details about $k$ and its influence on clustering performance are shown in Appendix~\ref{A5} of the supplementary material.

\begin{table}[!htbp]
\centering
\caption{Details of the six benchmark datasets (including three non-graph datasets and three graph datasets) in graph clustering experiments.}
\renewcommand{\arraystretch}{1}
\begin{tabular}{lccccc}
\toprule
   Dataset name   &Type     &\# Total samples  & \# Classes &    \# Dimension                  \\
\midrule
   USPS           &Image    & 9,298            & 10                  &256                    \\

   HHAR           &Record   & 10,299           & 6                   &561                  \\

   REUTERS-10K    &Text     & 10,000           & 4                   &2,000        \\

   ACM            &Graph    & 3,025            & 3                   &1,870           \\

   DBLP           &Graph    & 4,058            & 4                   &334       \\

   CITESEER       &Graph    & 3,327            & 6                   &3,703                         \\

\bottomrule

\end{tabular}
\label{Dataset2}
\end{table}
%

For fair comparison, we follow the settings in \citep{bo2020structural} to construct our model with a $d$-500-500-2000-$d'$ fully-connected encoder, the same size for GCN, and a symmetric decoder. The dimension of the latent layer $d'$ is fixed as 10.  We first pre-train an auto-encoder network for 30 epochs without GCN to initialize the model, then train the model for at least 200 epochs until it meets convergence. The learning rate is set to 1e-3 for USPS, HHAR, and ACM, 2e-4 for REUTERS, and 1e-4 for DBLP and CITESEER. We also conduct ablation study in Appendix~\ref{A4} to validate the effect of each component in our method. Moreover, we further discuss about the influence of two hyper-parameters $\gamma$ and $\beta$, the threshold of confident assignment $\delta$, and the number of nearest neighbors $k'$ on clustering performance in Appendix~\ref{A5}. We run the method for ten times and report the means and standard deviations of ACC, NMI and ARI.

\subsection{Comparative results}
Table~\ref{clustering_non_graph} and Table~\ref{clustering_graph} shows the clustering results of each method on USPS, REUTERS-10, DBLP, and CITESEER, while the results of HHAR and ACM are presented in Appendix~\ref{A3} for saving the space. First, we see that while the approaches using graph information (GAE and VGAE) perform well on natural graph-structured datasets such as CITESEER, they do not obtain good performance on non-graph-structured datasets such as USPS. This is mainly due to the fact that the graph information constructed on non-graph structured data is not sufficient to help capture the latent features of the data well. Second, deep clustering methods that introduce the clustering objective for the joint optimization of clustering and feature learning (DEC and IDEC) exhibit better performance compared to GAE and VGAE. More importantly, those methods combining graph neural networks and deep clustering (DAEGC, SDCN, and AGCN) show encouraging clustering performance, suggesting that considering the graph structure information of the data can benefit the label assignment in the clustering layer.
Third, our method obtains the top-two clustering performance on all tested datasets. Especially on the non-graph datasets such as USPS, the proposed method achieves 86.04\%, 83.19\%, 80.71\% with respect to ACC, NMI, and ARI, which outperforms state-of-the-art SDCN and AGCN with a large margin. This adequately demonstrates that the proposed method can capture informative clustering structures via minimizing the intra-cluster discrepancy and maximizing the inter-cluster discrepancy.

\begin{table*}[h]
\centering
\caption{Clustering performance on two non-graph data sets USPS and REUTERS-10K. The best result in each case is highlighted in bold.}
\resizebox{\textwidth}{!}{
\setlength{\tabcolsep}{1.2mm}
\renewcommand{\arraystretch}{1.3}
\begin{tabular}{l|ccc|cccc}

\toprule
\multirow{2}{*}{Methods/Datasets} & \multicolumn{3}{c|}{USPS} & \multicolumn{3}{c}{REUTERS-10K}  \\
\cmidrule(r){2-4} \cmidrule(r){5-7}
&  ACC     &  NMI &  ARI
&  ACC     &  NMI &  ARI     \\

\midrule

$k$-means~\citep{macqueen1967some}    &66.82$\pm$0.04 &62.63$\pm$0.05 &54.55$\pm$0.06    &54.04$\pm$0.01 &41.54$\pm$0.51 &27.95$\pm$0.38  \\


AE~\citep{hinton2006reducing}        &71.04$\pm$0.03 &67.53$\pm$0.03  &58.83$\pm$0.05     &74.90$\pm$0.21   &49.69$\pm$0.29   &49.55$\pm$0.37        \\

DEC~\citep{xie2016unsupervised}     &73.31$\pm$0.17 &70.58$\pm$0.25  &63.70$\pm$0.27     &73.58$\pm$0.13   &47.50$\pm$0.34   &48.44$\pm$0.14          \\

IDEC~\citep{guo2017improved}        &76.22$\pm$0.12 &75.56$\pm$0.06  &67.86$\pm$0.12     &75.43$\pm$0.14   &50.28$\pm$0.17   &51.26$\pm$0.21      \\

GAE~\citep{kipf2016variational}    &63.10$\pm$0.33 &60.69$\pm$0.58  &50.30$\pm$0.55     &54.40$\pm$0.27   &25.92$\pm$0.41   &19.61$\pm$0.22         \\

VGAE~\citep{kipf2016variational}  &56.19$\pm$0.72 &51.08$\pm$0.37  &40.96$\pm$0.59     &60.85$\pm$0.23   &25.51$\pm$0.22   &26.18$\pm$0.36         \\

DAEGC~\citep{wang2019attributed} &73.55$\pm$0.40 &71.12$\pm$0.24  &63.33$\pm$0.34     &65.50$\pm$0.13   &30.55$\pm$0.29   &31.12$\pm$0.18         \\

ARGA~\citep{pan2020learning}   &66.80$\pm$0.70 &61.60$\pm$0.30  &51.10$\pm$0.60     &56.20$\pm$0.20   &28.70$\pm$0.30   &24.50$\pm$0.40         \\

SDCN~\citep{bo2020structural}   &78.08$\pm$0.19 &79.51$\pm$0.27  &71.84$\pm$0.24     &77.15$\pm$0.21   &50.82$\pm$0.21   &55.36$\pm$0.37         \\

AGCN~\citep{peng2021attention}   &80.98$\pm$0.28 &79.64$\pm$0.32  &73.61$\pm$0.43     &79.30$\pm$1.07   &57.83$\pm$1.01   &60.55$\pm$1.78         \\

\midrule
DDAC-G    &\bf86.04$\pm$0.38 &\bf83.19$\pm$0.64  &\bf80.71$\pm$0.84     &\bf79.45$\pm$1.22   &\bf63.52$\pm$1.09   &\bf60.57$\pm$1.88         \\

\bottomrule

\end{tabular}}
\label{clustering_non_graph}
\end{table*}

\begin{table*}[h]
\centering
\caption{Clustering performance on two graph data sets DBLP and CITESEER. The best result in each case is highlighted in bold.}
\resizebox{\textwidth}{!}{
\setlength{\tabcolsep}{1.2mm}
\renewcommand{\arraystretch}{1.3}
\begin{tabular}{l|ccc|cccc}

\toprule
\multirow{2}{*}{Methods/Datasets} & \multicolumn{3}{c|}{DBLP} & \multicolumn{3}{c}{CITESEER}  \\
\cmidrule(r){2-4} \cmidrule(r){5-7}
&  ACC     &  NMI &  ARI
&  ACC     &  NMI &  ARI     \\

\midrule

$k$-means~\citep{macqueen1967some}    &38.65$\pm$0.65 &11.45$\pm$0.38 &6.97$\pm$0.39   &39.32$\pm$3.17 &16.94$\pm$3.22 &13.43$\pm$3.02  \\


AE~\citep{hinton2006reducing}        &51.43$\pm$0.35 &25.40$\pm$0.16  &12.21$\pm$0.43     &57.08$\pm$0.13 &27.64$\pm$0.08  &29.31$\pm$0.14        \\

DEC~\citep{xie2016unsupervised}     &58.16$\pm$0.56 &29.51$\pm$0.28  &23.92$\pm$0.39     &55.89$\pm$0.20   &28.34$\pm$0.30   &28.12$\pm$0.36          \\

IDEC~\citep{guo2017improved}        &60.31$\pm$0.62 &31.17$\pm$0.50  &25.37$\pm$0.60     &60.49$\pm$1.42   &27.17$\pm$2.40   &25.70$\pm$2.65      \\

GAE~\citep{kipf2016variational}    &61.21$\pm$1.22 &30.80$\pm$0.91  &22.02$\pm$1.40     &61.35$\pm$0.80   &34.63$\pm$0.65   &33.55$\pm$1.18         \\

VGAE~\citep{kipf2016variational}  &58.59$\pm$0.06 &26.92$\pm$0.06  &17.92$\pm$0.07     &60.97$\pm$0.36   &32.69$\pm$0.27   &33.13$\pm$0.53         \\

DAEGC~\citep{wang2019attributed} &62.05$\pm$0.48 &32.49$\pm$0.45  &21.03$\pm$0.52     &64.54$\pm$1.39   &36.41$\pm$0.86   &37.78$\pm$1.24        \\

ARGA~\citep{pan2020learning}   &61.60$\pm$1.00 &26.80$\pm$1.00  &22.70$\pm$0.30     &56.90$\pm$0.70   &34.50$\pm$0.80   &33.40$\pm$1.50         \\

SDCN~\citep{bo2020structural}   &68.05$\pm$1.81 &39.50$\pm$1.34  &39.15$\pm$2.01     &65.96$\pm$0.31   &38.71$\pm$0.32   &40.17$\pm$0.43         \\

AGCN~\citep{peng2021attention}   &73.26$\pm$0.37 &39.68$\pm$0.42  &42.49$\pm$0.31     &68.79$\pm$0.23   &41.54$\pm$0.30   &43.79$\pm$0.31        \\

\midrule
DDAC-G     &\bf75.26$\pm$1.01 &\bf48.80$\pm$0.72  &\bf49.46$\pm$1.70     &\bf69.94$\pm$0.85   &\bf43.96$\pm$0.82  &\bf46.79$\pm$0.71        \\

\bottomrule

\end{tabular}}
\label{clustering_graph}
\end{table*}

\begin{table}[!htbp]
	\centering
	\caption{The ablation studies of the proposed method on three datsets (USPS, HHAR, ACM), and the best results are marked in \textbf{Bold}. CA denotes the selection mechanism of confident assignment.}
		\begin{tabular}{l|ccc|ccc}
	\toprule

Datasets              &CA &$\mathcal{L}_{\textrm{orth}}$ &$\mathcal{L}_{\textrm{disc}}$ & ACC   &NMI   &ARI          \\ \midrule
\multirow{3}{*}{USPS}
                      &$\times$   &$\times$   &\checkmark     &79.69$\pm$0.75 &81.59$\pm$0.18 &74.11$\pm$0.16         \\
			          &$\times$   &\checkmark &\checkmark     &82.26$\pm$0.76 &80.86$\pm$0.71 &75.90$\pm$1.26         \\
			          &\checkmark &\checkmark &\checkmark     &\bf86.04$\pm$0.38   &\bf83.19$\pm$0.64    &\bf80.71$\pm$0.84       \\
\midrule
\multirow{3}{*}{HHAR}
                      &$\times$   &$\times$   &\checkmark     &86.14$\pm$1.87 &82.77$\pm$0.63 &75.72$\pm$1.33         \\
			          &$\times$   &\checkmark &\checkmark     &86.84$\pm$1.59 &83.04$\pm$0.52 &76.45$\pm$0.91          \\
			          &\checkmark &\checkmark &\checkmark     &\bf88.24$\pm$0.41 &\bf83.23$\pm$0.57 &\bf77.30$\pm$0.81         \\
\midrule
\multirow{3}{*}{ACM}
                      &$\times$   &$\times$   &\checkmark     &90.57$\pm$0.42 &68.40$\pm$0.39 &74.18$\pm$0.79        \\
			          &$\times$   &\checkmark &\checkmark     &90.61$\pm$0.28 &\bf68.84$\pm$0.70 &74.21$\pm$0.71        \\
			          &\checkmark &\checkmark &\checkmark     &\bf90.74$\pm$0.30 &68.61$\pm$0.71 &\bf74.52$\pm$0.83          \\

 \bottomrule
	\end{tabular}
	\label{tab:ablationstudy}
\end{table}

\subsection{Ablation study}
We use USPS, HHAR and ACM to show the effectiveness of each component in the proposed DDAC-G. The experimental results are illustrated in Shown in Table~\ref{tab:ablationstudy}, the clustering performance in almost all cases (datasets and metrics) decreases when one component of DDAC-G is ablated. Besides, the selection mechanism of confident assignment works very well on USPS and HHAR, but has limited improvement on ACM, which is probably due to the small number of categories and highly informative graph information in ACM.

\section{Conclusion}
\label{sec_conclusion}
In this work, we have introduced two novel clustering methods DDAC and DDAC-G. The main idea is to obtain small intra-cluster discrepancy and large inter-cluster discrepancy, thereby project the data into a cluster-informative low-dimensional feature space, in which the clusters can be easily identified.  The method can be regarded as an extension of LDA to unsupervised learning and deep learning.  Extensive experiments demonstrated the effectiveness of the proposed methods. One limitation of DDAC-G as well as other graph based clustering methods is that they have high computational costs on very large datasets owing to the $n\times n$ similarity matrix, which may be considered in future study.


\bibliographystyle{named}
\bibliography{nips22}

\appendix

\section{Appendix}

\subsection{More details about the datasets}
\label{A1}

\textbf{MNIST}\footnote{http://yann.lecun.com/exdb/mnist/} is a well-known handwritten digit dataset for machine learning. It contains handwritten font images from 0 to 9, with 60,000 training samples and 10,000 test samples. Each sample is a grey-scale image of size 28 $\times$ 28.

\textbf{Fashion-MNIST}\footnote{https://www.kaggle.com/zalando-research/fashionmnist} contains 10 classes of fashion style images such as bag, coat, and dress, etc. It shares the same composition as MNIST (60,000 samples for training and 10,000 samples for test), as well as the same image size (28 $\times$28) of each sample.

\textbf{USPS:} USPS\footnote{https://www.csie.ntu.edu.tw/~cjlin/libsvmtools/datasets/} is a grey-scale digit image dataset that consists of 9,298 images in total. It contain 10 categories from the digits 0 to 9, and each sample is a gray-scale image of size 16 $\times$ 16.

\textbf{REUTERS-10K}\footnote{https://keras.io/api/datasets/reuters/} is a subset of the REUTERS large text dataset, and we utilize 10,000 samples containing 4 categories (corporate/industrial, economics, government/social, and markets) for our experiments in this paper. Each sample is a 2,000-dimensional feature vector.

\textbf{HHAR}\footnote{https://archive.ics.uci.edu/ml/datasets/Human+Activity+Recognition+Using+Smartphones} (Heterogeneity Human Activity Recognition) contains 10,299 samples with 6 categories of human activities, including walking, sitting, standing, walking-upstairs, walking-downstairs, and laying while wearing a smartphone. Each of the records is described as a 561-dimensional feature vector.

\textbf{DBLP}\footnote{https://dblp.uni-trier.de} is a dataset composed of a relationship network of authors, where two authors have an edge between them if they are co-authors. It consists of 4,058 samples in 4 domains (database, data mining, information retrieval and machine learning). Each sample is characterised by 334 features.

\textbf{CITESEER}\footnote{http://citeseerx.ist.psu.edu/index} is a citation network dataset containing 3,327 samples from 6 domains (agencies, artificial intelligence, database, human-computer interaction, information retrieval, and machine language). Each sample is represented by a 3,703-dimensional feature vector.

\textbf{ACM}\footnote{http://dl.acm.org/} is a dataset consisting of papers published in ACM, which contains 3,025 samples and is divided into 3 categories by research area (database, data mining, and wireless communication). Each sample is represented by a 1,870-dimensional feature vector.

Table~\ref{Dataset} summaries the details of the eight benchmark datasets.
\begin{table}[!htbp]
\centering
\caption{Details of the eight benchmark datasets in our experiments.}
\begin{tabular}{lccccc}
\toprule
   Dataset name   &Type     &\# of samples  & \# of classes &    \# of features                  \\
\midrule
   MNIST          &Image    & 70,000           & 10                  &784                    \\

   Fashion-MNIST  &Image    & 70,000           & 10                  &784                    \\

   USPS           &Image    & 9,298            & 10                  &256                    \\

   HHAR           &Record   & 10,299           & 6                   &561                  \\

   REUTERS-10K    &Text     & 10,000           & 4                   &2,000        \\

   ACM            &Graph    & 3,025            & 3                   &1,870           \\

   DBLP           &Graph    & 4,058            & 4                   &334       \\

   CITESEER       &Graph    & 3,327            & 6                   &3,703                         \\

\bottomrule

\end{tabular}
\label{Dataset}
\end{table}

\subsection{Details of the network architecture of DDAC}
\label{A2}
We run all experiments on NVIDIA RTX3080 GPU with 32GB RAM, CUDA 11.0 and cuDNN 8.0. We construct the DDAC model with FCN and CNN, yielding DDAC-FCN and DDAC-CNN respectively. For the DDAC-FCN, we use the following network settings (detailed in Table~\ref{FCN-based-net}). For MNIST, we construct the network with a $d$-500-500-1000-$d'$ fully-connected encoder and a symmetric decoder, where the latent dimension $d'$ is 10. For Fashion-MNIST, we construct the network with a $d$-500-500-2000-$d'$ fully-connected encoder and a symmetric decoder. For the CNN-based version, we utilize the same network architecture for MNIST and Fashion-MNIST, and the technical details of the network architecture are described in Table~\ref{CNN-based-net}. Besides, we run $k$-means 20 times with the pre-trained features to initialize the clustering centers, and set $\beta = 0.01, \gamma=$ 1e-5 for MNIST, while $\beta = 0.005, \gamma = $ 1e-3 for Fashion-MNIST.

\begin{table}[!htbp]
\centering
\caption{Network architecture of the proposed DDAC method constructed with FCN.}
\label{FCN-based-net}
\begin{tabular}{|c|l|l|}
\hline
Dataset                        & Encoder                   & Decoder                  \\ \hline
\multirow{4}{*}{MNIST}         & Linear(784, 500), ReLU()  & Linear(10, 1000),ReLU()  \\
                               & Linear(500, 500), ReLU()  & Linear(1000, 500),ReLU() \\
                               & Linear(500, 1000), ReLU() & Linear(500, 500),ReLU()  \\
                               & Linear(1000, 10)          & Linear(500, 784)         \\ \hline
\multirow{4}{*}{Fashion-MNIST} & Linear(784, 500), ReLU()  & Linear(10, 2000),ReLU()  \\
                               & Linear(500, 500), ReLU()  & Linear(2000, 500),ReLU() \\
                               & Linear(500, 2000), ReLU() & Linear(500, 500),ReLU()  \\
                               & Linear(2000, 10)          & Linear(500, 784)         \\ \hline
\end{tabular}
\end{table}

\begin{table}[!htbp]
\caption{Network architecture of the proposed DDAC method constructed with CNN.}
\label{CNN-based-net}
\begin{tabular}{|l|}
\hline
Ecnoder                                                                                       \\ \hline
Conv2d(in\_channel=1, out\_channel=16, kernel\_size=3, stride=1, padding=1)                   \\
BatchNorm2d(16), ReLU()                                                                       \\
Conv2d(in\_channel=16, out\_channel=32, kernel\_size=3, stride=2, padding=1)                  \\
BatchNorm2d(32), ReLU()                                                                       \\
Conv2d(in\_channel=32, out\_channel=32, kernel\_size=3, stride=1, padding=1)                  \\
BatchNorm2d(32), ReLU()                                                                       \\
Conv2d(in\_channel=32, out\_channel=16, kernel\_size=3, stride=2, padding=1)                  \\
BatchNorm2d(16), ReLU(), Flatten()                                                            \\
Linear(784, 256), ReLU()                                                                      \\
Linear(256, 10)                                                                               \\ \hline
Decoder                                                                                       \\ \hline
Linear(10, 256), ReLU()                                                                       \\
Linear(256, 784)                                                                              \\
Reshape(16, 3, 3)                                                                             \\
ConvTranspose2d(in\_channel=16, out\_channel=32, kernel\_size=3, stride=2, output\_padding=1) \\
BatchNorm2d(32), ReLU()                                                                       \\
ConvTranspose2d(in\_channel=32, out\_channel=32, kernel\_size=3, stride=1, padding=1)         \\
BatchNorm2d(32), ReLU()                                                                       \\
ConvTranspose2d(in\_channel=32, out\_channel=16, kernel\_size=3, stride=2, output\_padding=1) \\
BatchNorm2d(16), ReLU()                                                                       \\
ConvTranspose2d(in\_channel=16, out\_channel=1, kernel\_size=3, stride=1, padding=1)          \\ \hline
\end{tabular}
\end{table}

\begin{figure}[!htbp]
\centering
\includegraphics[width=4.5in]{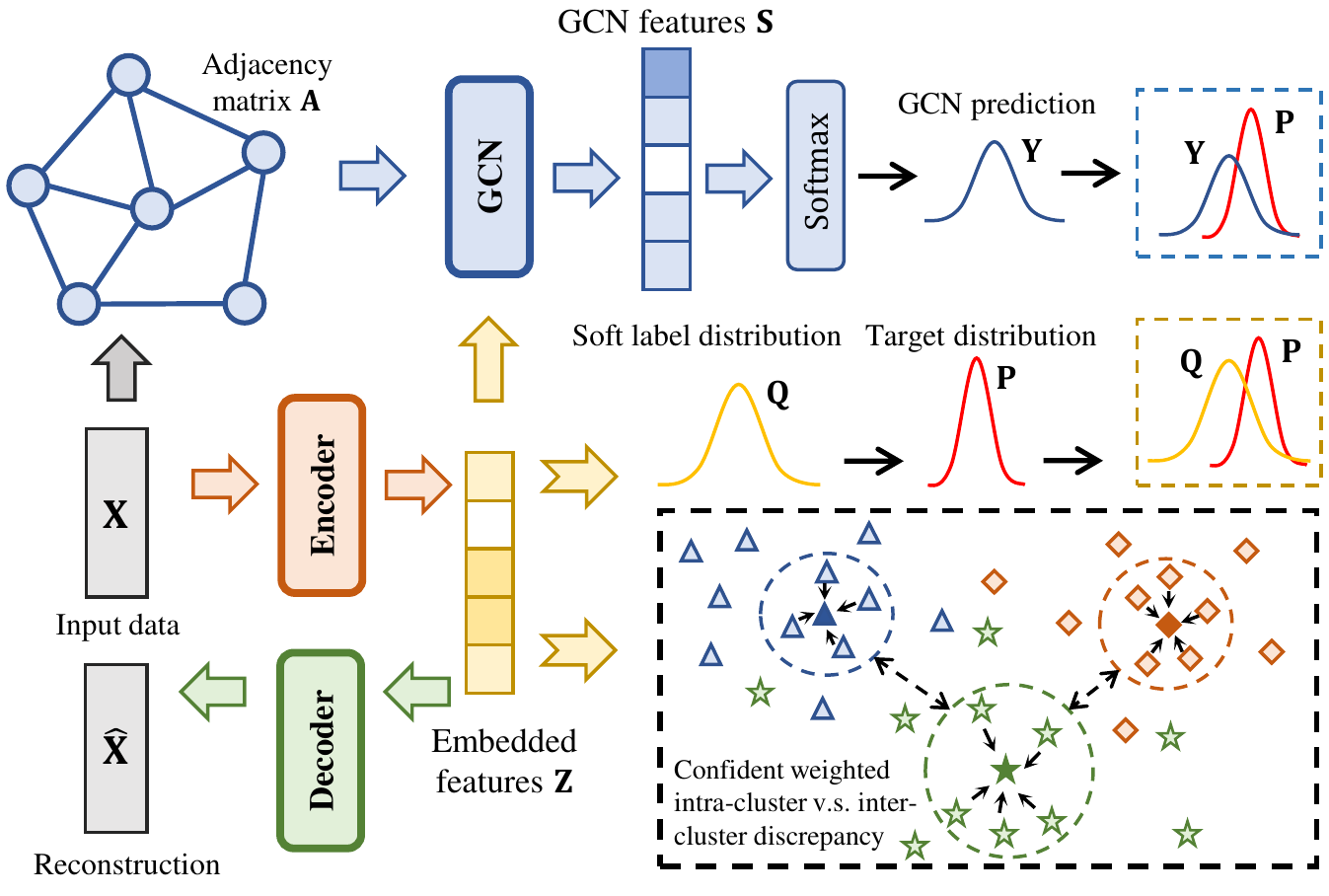}\\
\caption{The network architecture of the proposed method DDAC-G. The original data $\mathbf{X}$ and adjacency matrix $\mathbf{A}$ serve as the input to the first GCN layer, and then the embedded features $\mathbf{S}$ and $\mathbf{A}$ serve as the input to each subsequent layer of GCN. The networks are jointly optimized via minimizing the data reconstruction loss, KL divergences between $\mathbf{Q}$ and $\mathbf{P}$ and between $\mathbf{Y}$ and $\mathbf{P}$, and the objective of confident weighted intra-cluster v.s.  inter-cluster discrepancy.}
\label{DDGCNetwork}
\end{figure}

The network architecture of the extension of our method for graph clustering, i.e., deep discriminative graph clustering (DDAC-G), is presented in Figure~\ref{DDGCNetwork}. The algorithm of DDAC-G is illustrated in Algorithm~\ref{AlgorithmDDGC}. We follow the network architecture that used in \citep{bo2020structural} and~\citep{peng2021attention} to guarantee fair comparison. The network settings are presented in Table~\ref{DDGC-net}.

\begin{algorithm}[h]
\caption{Deep Discriminative Graph Clustering (DDAC-G)}
\label{AlgorithmDDGC}
\textbf{Input}: Data matrix $\mathbf{X}$, adjacency matrix $\mathbf{A}$, dimension of latent space $d'$, number of clusters $k$, learning rate $\rho$, hyper-parameters $\beta$ and $\gamma$, threshold of confident assignment $\delta$, total training iterations $T_{iter}$.\\
\textbf{Output}: Cluster labels $C$.

\begin{algorithmic}[1]
\STATE {Initialize the network parameters $\Theta= \left \{ \theta, \theta' \right \}$ of AE with the pre-trained model;}
\STATE {Initialize the cluster centroids $\bm{\mu}$ with $k$-means;}
      \FOR{$iter=1$ to $T_{iter}$}
      \STATE {Obtain the learned latent representation $\mathbf{Z}$ of AE;}
      \STATE {Compute the soft label distribution $\mathbf{Q}$ by Eq.~\eqref{eqsoft};}
      \STATE {Compute the target distribution $\mathbf{P}$ by Eq.~\eqref{eqtarget};}
      \STATE {Compute the prediction $\mathbf{Y}$ of GCN by Eq.~\eqref{eqgcnzoutput};}
      \STATE {Obtain the selective index set $\mathbf{t}$ by Eq.~\eqref{eqthreshold};}
      \STATE {Calculate $\mathcal{L}_{\textrm{recon}}$, $\mathcal{L}_{\textrm{clus}}$, $\mathcal{L}_{\textrm{disc}}$, and $\mathcal{L}_{\textrm{orth}}$ through Eq.~\eqref{eq:ReconstructionLoss},~\eqref{eqlc},~\eqref{eqdisc},~\eqref{eqorth} respectively;}
      \STATE {Update the network parameters $\Theta$, $ \left \{ \bm{\mu} \right \}$, and $\mathcal{W}$ by minimizing Eq~\eqref{eq_all};}
   \ENDFOR
\STATE {Calculate the final cluster results as $c_{i} = \arg\max y_{ij}$.}
\RETURN {$C$.}
\end{algorithmic}
\end{algorithm}

\begin{table}[!htbp]
\centering
\caption{Network architecture of the extension of the proposed method (DDAC-G) for graph clustering.}
\label{DDGC-net}
\begin{tabular}{|c|l|l|}
\hline
Network                            & Encoder                   & Decoder                  \\ \hline
\multirow{4}{*}{AE}            & Linear(784, 500), ReLU()  & Linear(10, 2000),ReLU()  \\
                               & Linear(500, 500), ReLU()  & Linear(2000, 500),ReLU() \\
                               & Linear(500, 2000), ReLU() & Linear(500, 500),ReLU()  \\
                               & Linear(2000, 10)          & Linear(500, 784)         \\ \hline

Network                        & GCN                   &                   \\ \hline

\multirow{5}{*}{GNN}           & GCN\_Layer(784, 500), ReLU()     &           \\
                               & GCN\_Layer(500, 500), ReLU()     &                  \\
                               & GCN\_Layer(500, 2000), ReLU()    &                              \\
                               & GCN\_Layer(2000, 10) , ReLU()    &                               \\
                               & GCN\_Layer(10, 10)   &   \\\hline
\end{tabular}
\end{table}

\subsection{More experimental results}
\label{A3}
\begin{figure}[!htbp]
\centering
\includegraphics[width=5.4in]{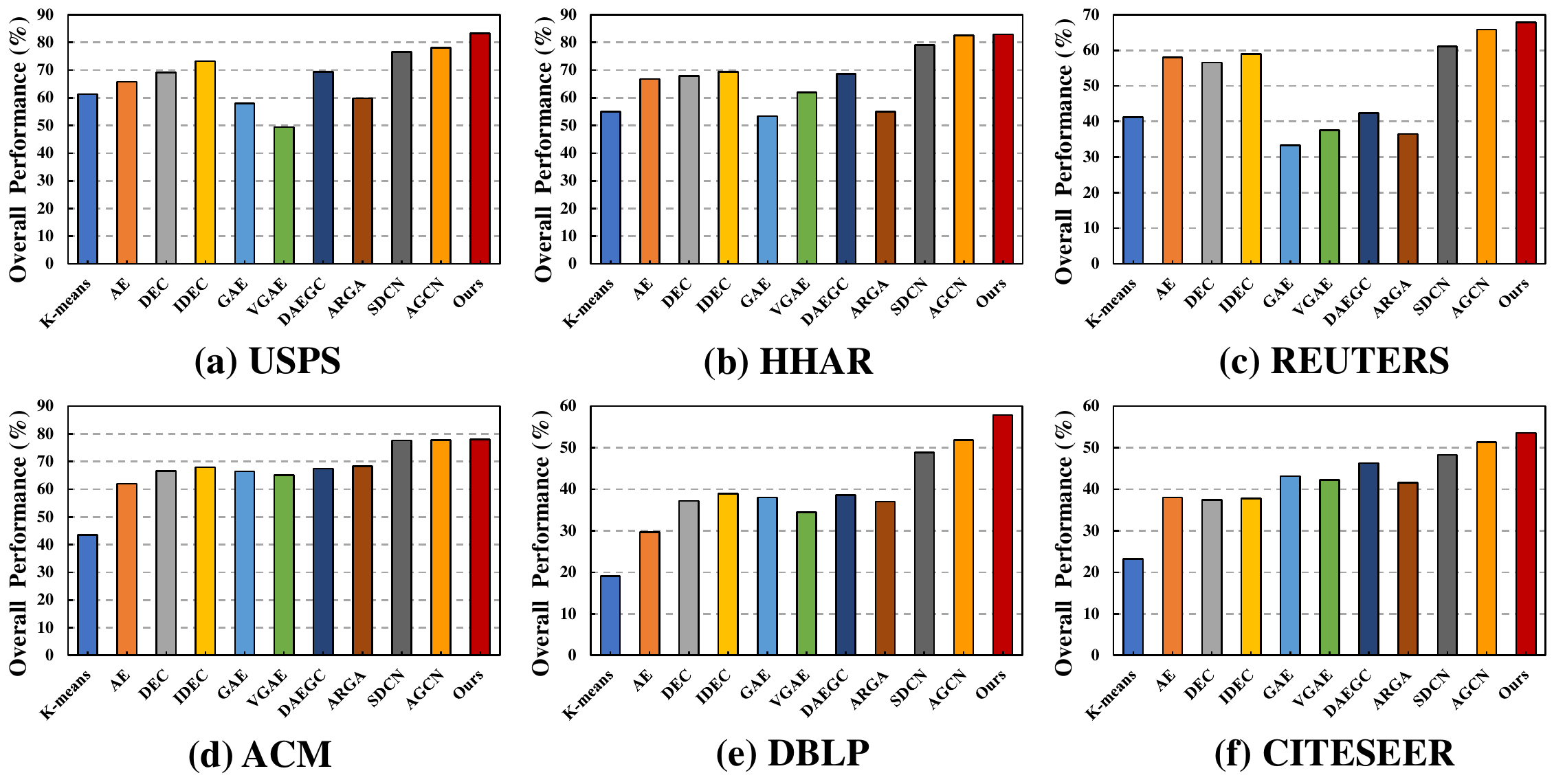}\\
\caption{Comparison of each clustering method in terms of overall clustering performance. The proposed method achieves the best results on five datasets and is also comparable to the best performing DFCN on ACM.}
\label{OverallComparison}
\end{figure}

We further conduct experiment on a non-graph dataset HHAR and a graph dataset ACM, and the results are reported in Table~\ref{clustering_results}. From this table, we see that the proposed DDAC-G method outperform most baseline methods with large margin. Besides, DDAC-G achieves 0.13\%, 0.79\%, 0.23\% improvement in ACC, NMI and ARI on HHAR, and 0.15\%, 0.23\%, 0.32\% improvement on ACM compared to state-of-the-art AGCN method. We also show the average scores of the three metrics in Figure~\ref{OverallComparison} to provide an overall evaluation.

\begin{table*}[!htbp]
\centering
\caption{Clustering performance on the non-graph dataset HHAR and the graph dataset ACM. Note that the results marked in $\textbf{bold}$ denote the best clustering performance.}
\resizebox{\textwidth}{!}{
\setlength{\tabcolsep}{1.0mm}
\renewcommand{\arraystretch}{1.3}
\begin{tabular}{l|ccc|cccc}

\toprule
\multirow{2}{*}{Methods/Data sets} & \multicolumn{3}{c|}{HHAR (non-graph)} & \multicolumn{3}{c}{ACM (graph)}  \\
\cmidrule(r){2-4} \cmidrule(r){5-7}
&  ACC     &  NMI &  ARI
&  ACC     &  NMI &  ARI     \\

\midrule

$k$-means~\citep{macqueen1967some}    &59.98$\pm$0.02  &58.86$\pm$0.01  &46.09$\pm$0.02      &67.31$\pm$0.71  &32.44$\pm$0.46  &30.60$\pm$0.69  \\

AE~\citep{hinton2006reducing}      &68.69$\pm$0.31  &71.42$\pm$0.97  &60.36$\pm$0.88     &81.83$\pm$0.08  &49.30$\pm$0.16  &54.64$\pm$0.16        \\

DEC~\citep{xie2016unsupervised}    &69.39$\pm$0.25  &72.91$\pm$0.39  &61.25$\pm$0.51      &84.33$\pm$0.76  &54.54$\pm$1.51  &60.64$\pm$1.87         \\

IDEC~\citep{guo2017improved}        &71.05$\pm$0.36  &74.19$\pm$0.39  &62.83$\pm$0.45      &85.12$\pm$0.52  &56.61$\pm$1.16  &62.16$\pm$1.50   \\

GAE~\citep{kipf2016variational}     &62.33$\pm$1.01  &55.06$\pm$1.39  &42.63$\pm$1.63      &84.52$\pm$1.44  &55.38$\pm$1.92  &59.46$\pm$3.10        \\

VGAE~\citep{kipf2016variational}  &71.30$\pm$0.36  &62.95$\pm$0.36  &51.47$\pm$0.73      &84.13$\pm$0.22  &53.20$\pm$0.52  &57.72$\pm$0.67         \\

DAEGC~\citep{wang2019attributed}   &76.51$\pm$2.19  &69.10$\pm$2.28  &60.38$\pm$2.15      &86.94$\pm$2.83  &56.18$\pm$4.15  &59.35$\pm$3.89      \\

ARGA~\citep{pan2020learning}     &63.30$\pm$0.80  &57.10$\pm$1.40  &44.70$\pm$1.00      &86.10$\pm$1.20  &55.70$\pm$1.40  &62.90$\pm$2.10         \\

SDCN~\citep{bo2020structural}     &84.26$\pm$0.17  &79.90$\pm$0.09  &72.84$\pm$0.09      &90.45$\pm$0.18 &68.31$\pm$0.25  &73.91$\pm$0.40        \\

AGCN~\citep{peng2021attention}    &88.11$\pm$0.43  &82.44$\pm$0.62  &77.07$\pm$0.66      &90.59$\pm$0.15  &68.38$\pm$0.45  &74.20$\pm$0.38         \\

\midrule
DDAC-G       &\bf88.24$\pm$0.41  &\bf83.23$\pm$0.57  &\bf77.30$\pm$0.81      &\bf90.74$\pm$0.30 &\bf68.61$\pm$0.71  &\bf74.52$\pm$0.83        \\

\bottomrule

\end{tabular}}
\label{clustering_results}
\end{table*}

\begin{figure}[!htbp]
\centering
\includegraphics[width=5.4in]{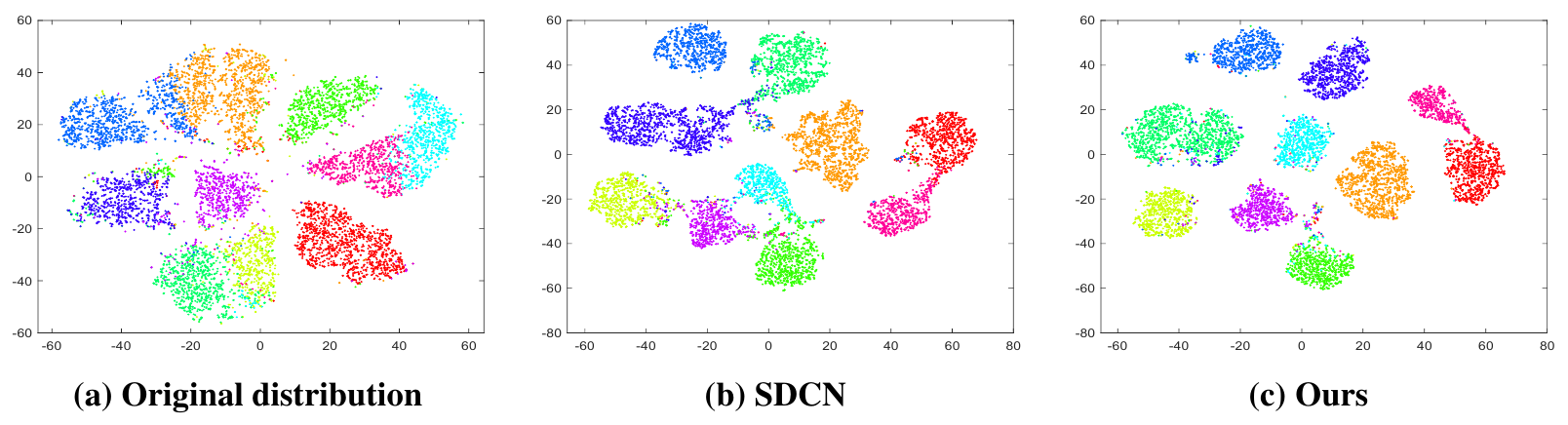}\\
\caption{The t-SNE visualization of the learned representations from SDCN and our method on USPS dataset, and the original distribution serves as the baseline.}
\label{visualization}
\end{figure}

To provide an intuitive comparison on clustering, we use t-SNE~\citep{van2008visualizing} to visualize the clustering performance on the USPS dataset. Figure~\ref{visualization} plots the visual comparison of raw data and the learned representations of SDCN and our method. We can observe that both SDCN and our method reveal better clustering structure compared to the original data distribution. Moreover, our method exhibits more compact intra-cluster structure, and more significant inter-cluster separation compared to SDCN, which implies that our method has  learnt a more discriminative representation.

\subsection{Parameter analysis}
\label{A5}
\begin{figure}[!htbp]
\centering
\includegraphics[width=5.4in]{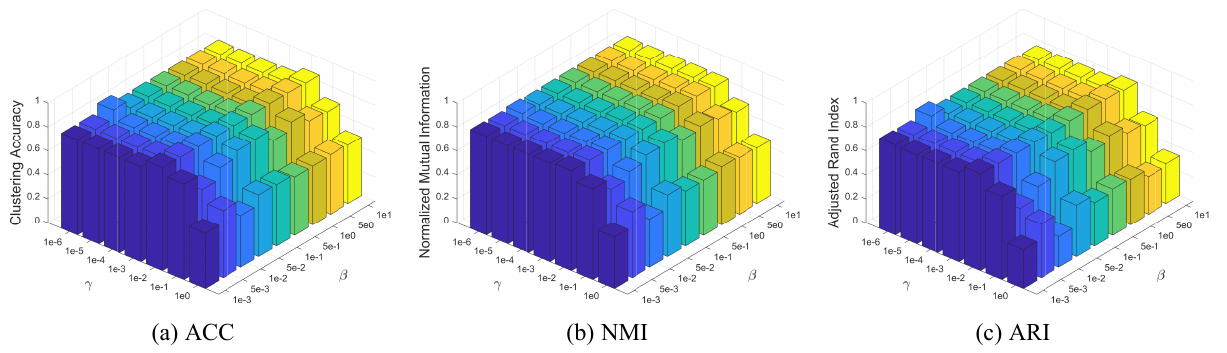}\\
\caption{The sensitivity analysis of two hyper-parameters $\beta$ and $\gamma$ on USPS dataset. Note that the values of $\beta$ and $\gamma$ range from ${\left \{ \textrm{1e-3}, \cdots, \textrm{1e1}\right \} }$ and $\left \{ \textrm{1e-6}, \cdots, \textrm{1e0} \right \} $, respectively.}
\label{Parameter}
\end{figure}

\paragraph{Analysis of hyper-parameters $\beta$ and $\gamma$.}
We conduct the sensitivity analysis of two main hyper-parameters in our method, i.e., $\beta$ and $\gamma$ that control the contribution of the $\mathcal{L}_{\mathrm{disc}}$ and $\mathcal{L}_{\mathrm{orth}}$ in optimization. Specifically, we select USPS as the tested dataset, vary the values of $\beta$ and $\gamma$ from ${\left \{ \textrm{1e-3}, \cdots, \textrm{1e1} \right \} }$ and $\left \{ \textrm{1e-6}, \cdots, \textrm{1e0} \right \} $, and show the clustering performance in Figure~\ref{Parameter}. It can be seen that the our method has the best clustering performance when $\beta = \textrm{1e-3}$ and $\gamma = \textrm{1e-2}$, which indicates the effectiveness of these two terms. However, either excessive $\beta$ or $\gamma$ can negatively affect the clustering performance due to their overemphasis in the optimization. Therefore, we empirically give the recommended range of values for $\beta$ and $\gamma$ as from 1e-3 to 1e-1 and from 1e-5 to 1e-2, respectively.

\begin{figure}[!htbp]
\centering
\includegraphics[width=5.4in]{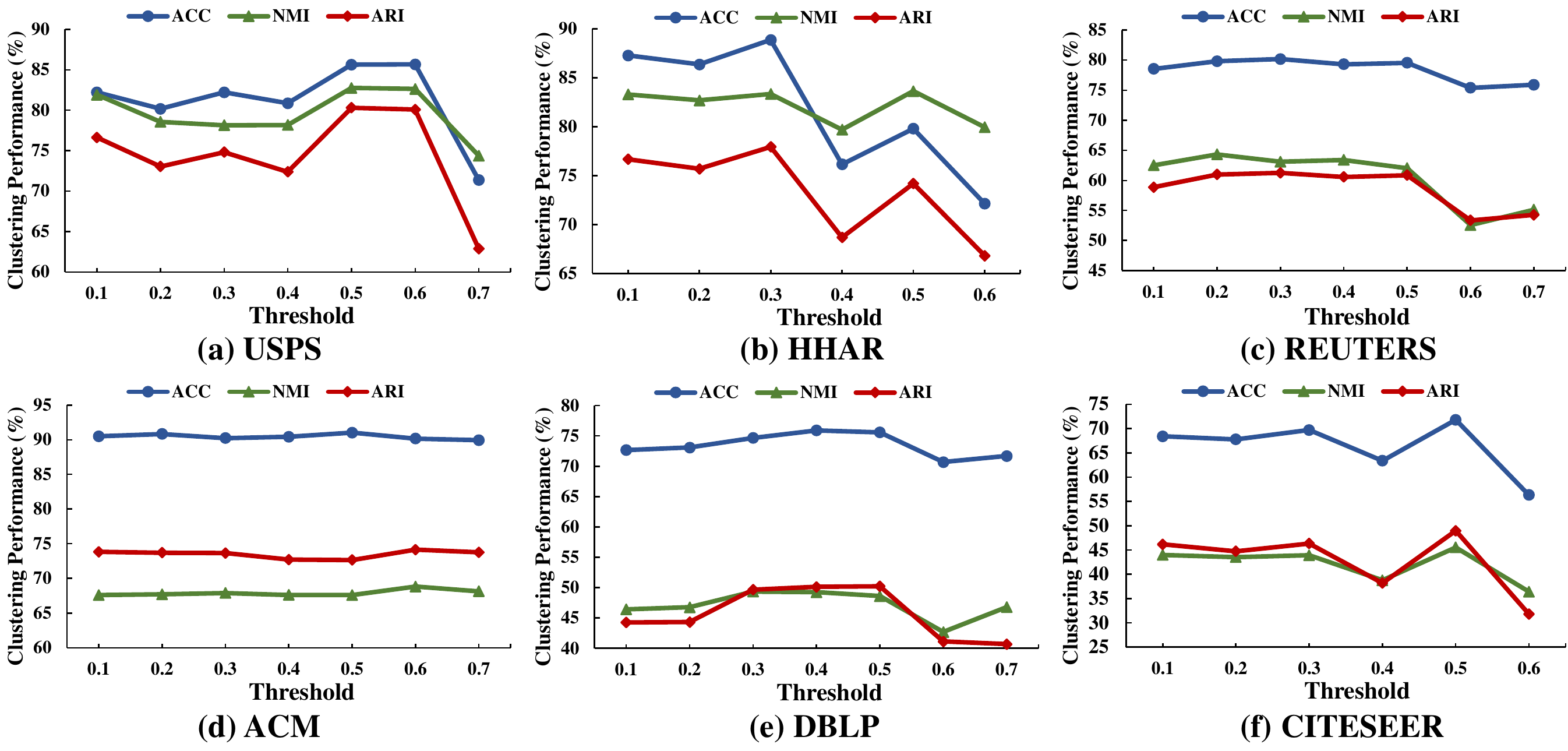}\\
\caption{The performance comparison under different confident assignment thresholds, note that we set the threshold to $\left \{0.1, \cdots, 0.6\right \} $ for HHAR and CITESEER since they fail to select eligible assignments when the threshold is set to 0.7.}
\label{DifferentRatio}
\end{figure}

\paragraph{Analysis of the confident assignment threshold.}
Figure~\ref{DifferentRatio} shows the clustering performance of our method under different confident assignment thresholds. We set the threshold to $\left \{0.1, \cdots, 0.7\right \} $ for USPS, REUTERS, ACM, and DBLP, and $\left \{0.1, \cdots, 0.6\right \} $ for HHAR and CITESEER since they fail to select eligible assignments when the threshold is set to 0.7. We can observe that filtering out some unconfident assignments by setting a threshold contributes to the clustering. Nevertheless, the clustering performance is negatively influenced if the threshold is set too high, because an excessively high threshold leads to select very few samples, which do not provide enough information of clusters. Furthermore, we can also find that the clustering performance of the ACM is only slightly affected by the threshold, which indicates that the graph of the ACM already provides good structural information available for the clustering task.

\begin{figure}[h]
\centering
\includegraphics[width=5.4in]{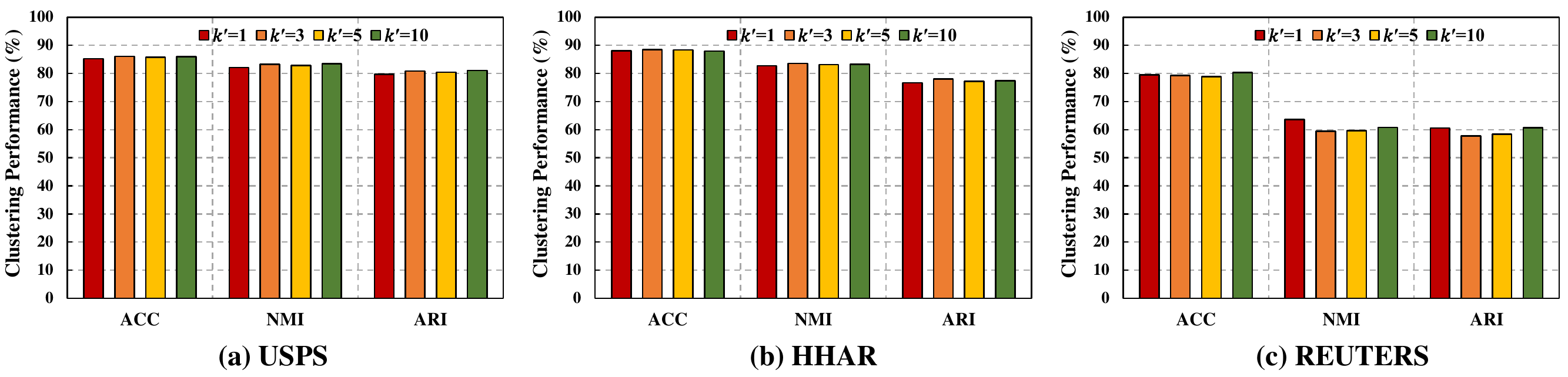}\\
\caption{Clustering performance variation with different number of nearest neighbors $k'$ in non-graph datasets. Note that $k'$ is set to $\left \{1, 3, 5, 10\right \}$, respectively.}
\label{DifferentK}
\end{figure}

\paragraph{Analysis of the number of nearest neighbors.}
Since we use  $k$NN to construct graphs for non-graph datasets, the clustering performance may be influenced by the value of $k$. Therefore, we set different number of nearest neighbors $k'$ in $k$NN to evaluate the influence of their graphs constructed for non-graph data on the clustering performance. Specifically, we set $k'$ to $\left \{1, 3, 5, 10\right \} $ and present the clustering results in Figure~\ref{DifferentK}. We can observe that the proposed DDAC-G method maintains stable clustering performance for different values of $k'$, which demonstrates the robustness of the proposed method with regard to the variation of $k'$. Nevertheless, it still achieves better performance at some specific $k'$ values, such as setting $k'$ to 3 for USPS and HHAR, and 1 for REUTERS.

\end{document}